\documentclass[conference]{IEEEtran}
\IEEEoverridecommandlockouts
\usepackage{cite}
\usepackage{amsmath,amssymb,amsfonts}
\usepackage{algorithmic}
\usepackage{graphicx}
\usepackage{textcomp}
\usepackage{xcolor}
\usepackage{array}
\usepackage[]{footmisc}
\usepackage{tikz}

\def\BibTeX{{\rm B\kern-.05em{\sc i\kern-.025em b}\kern-.08em
    T\kern-.1667em\lower.7ex\hbox{E}\kern-.125emX}}

\newcommand\copyrighttext{%
  \footnotesize \textcopyright 2021 IEEE. Personal use of this material is permitted.
  Permission from IEEE must be obtained for all other uses, in any current or future
  media, including reprinting/republishing this material for advertising or promotional
  purposes, creating new collective works, for resale or redistribution to servers or
  lists, or reuse of any copyrighted component of this work in other works.
}
\newcommand\copyrightnotice{%
\begin{tikzpicture}[remember picture,overlay]
\node[anchor=south,yshift=10pt] at (current page.south) {\fbox{\parbox{\dimexpr\textwidth-\fboxsep-\fboxrule\relax}{\copyrighttext}}};
\end{tikzpicture}%
}

\begin{document}

\title{Visual Question Answering based on Formal Logic}

\author{
\IEEEauthorblockN{Muralikrishnna G. Sethuraman}
\IEEEauthorblockA{\textit{School of Electrical and Computer Engineering} \\
\textit{Georgia Institute of Technology}\\
Atlanta, USA \\
muralikgs@gatech.edu}
\and
\IEEEauthorblockN{Ali Payani}
\IEEEauthorblockA{\textit{Cisco} \\
California, USA \\
apayani@cisco.com}
\and
\IEEEauthorblockN{Faramarz Fekri}
\IEEEauthorblockA{\textit{School of Electrical and Computer Engineering} \\
\textit{Georgia Institute of Technology}\\
Atlanta, USA \\
fekri@ece.gatech.edu}
\and
\IEEEauthorblockN{J. Clayton Kerce}
\IEEEauthorblockA{\centerline{\textit{Georgia Tech Research Institute}} \\
\textit{Georgia Institute of Technology}\\
Atlanta, USA \\
clayton.kerce@gtri.gatech.edu}    
}

\maketitle
\copyrightnotice

\begin{abstract}
  Visual question answering (VQA) has been gaining a lot of traction in the machine learning community in the recent years due to the challenges posed in understanding information coming from multiple modalities (i.e., images, language). In VQA, a series of questions are posed based on a set of images and the task at hand is to arrive at the answer. To achieve this, we take a symbolic reasoning based approach using the framework of formal logic. The image and the questions are converted into symbolic representations on which explicit reasoning is performed. We propose a formal logic framework where (i) images are converted to logical background facts with the help of scene graphs, (ii) the questions are translated to first-order predicate logic clauses using a transformer based deep learning model, and (iii) perform satisfiability checks, by using the background knowledge and the grounding of predicate clauses, to obtain the answer. Our proposed method is highly interpretable and each step in the pipeline can be easily analyzed by a human. We validate our approach on the CLEVR and the GQA dataset. We achieve near perfect accuracy of 99.6\% on the CLEVR dataset comparable to the state of art models, showcasing that formal logic is a viable tool to tackle visual question answering. Our model is also data efficient, achieving 99.1\% accuracy on CLEVR dataset when trained on just 10\% of the training data.

\end{abstract}

\begin{IEEEkeywords}
Visual Question Answering, formal logic, transformers, interpretable learning
\end{IEEEkeywords}

\section{Introduction.}
In the recent years AI research has been moving towards
solving increasingly difficult problems with the goal of building a general purpose intelligence system. To achieve this, these AI systems require an understanding of information
acquired through multiple modalities (visual, audio, language
etc.,) in order to make decisions related to various tasks. Visual
question answering (VQA) \cite{vqa_antol} is one such task that integrates
the domains of Computer Vision (CV) and Natural Language
Processing (NLP). 

\begin{figure}[t!]
\centering
\includegraphics[width=0.8\linewidth]{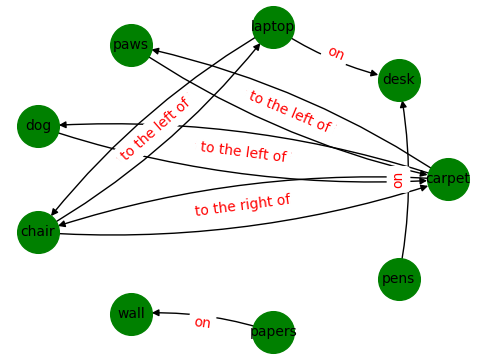}
\caption{Example of a scene graph taken from the GQA dataset \cite{gqa}.}
\label{fig:scene_graph}
\end{figure}

\begin{figure*}[t]
\centering
\includegraphics[width=\textwidth]{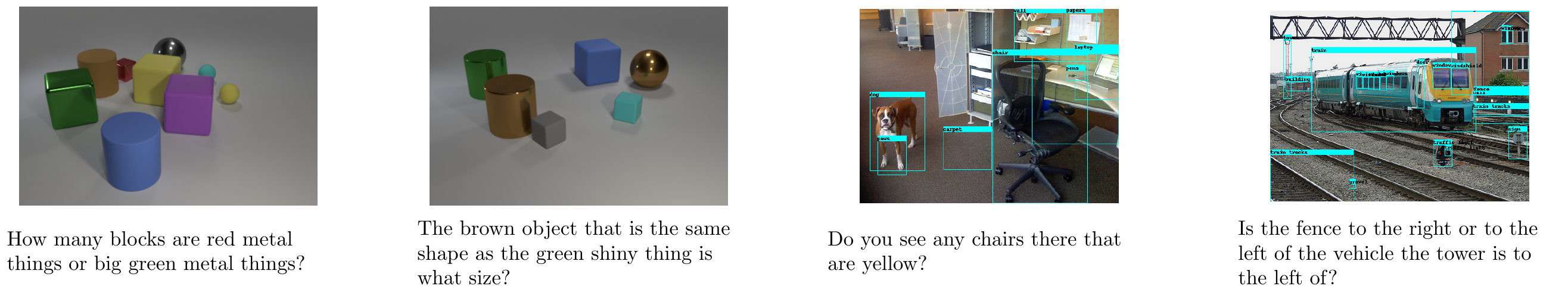}
\caption{Samples of images and its associated questions. The first two images from the left are taken from CLEVR \cite{clevr} dataset where all the images are rendered and only contains objects of simple shapes. The last two examples are taken from the GQA \cite{gqa} dataset containing images taken from natural settings.}
\label{fig:samples}
\end{figure*}

In VQA, a series of questions is posed based on a scene (image) (Figure \ref{fig:samples}) and the objective is to answer these questions based on the information provided. In the most general case, images can contain a lot of objects making it hard to parse the scene. In some cases the necessary information may not be readily available from the images and may depend some commonsense facts \cite{survey}. Additionally, questions posed could be of arbitrary complexity due to convoluted chains of logical reasoning. Moreover, models which directly learn input to output mappings are found to have greater sensitivity towards dataset biases, and fail to learn any underlying reasoning mechanisms \cite{biases}. These challenges make VQA a difficult problem to solve.

Nonetheless, there have been several attempts to convert images into a structured format, to aid the process of reasoning. Notably, scene graphs (Figure \ref{fig:scene_graph}) provide a structured representation of the objects in the scene along with the associated relations in the form of a graph. Here, the vertices of the graph correspond to the objects and the edges depict the relations between the objects. Significant contributions have been made towards converting images to scene graphs \cite{img2sg1, img2sg2}. Since a scene graph is a symbolic representation of a scene, this opens up the possibility of performing symbolic reasoning on images. Building on this, \cite{sgImgCap} show how scene graphs can be utilized for image captioning and \cite{sgVqa2} demonstrate the use of scene graphs in their VQA pipeline. In particular, \cite{sgVqa2} make use of Graph Neural Networks on top of the scene graph to create higher level abstractions, making it not very easily trainable. In this paper we directly operate on the scene graph instead of dealing with the images, but do so in the context of formal logic. The scene graphs are converted into a set of background facts about the scene which are easy to interpret and the translation requires very little effort.

In addition to the works on scene graphs, several attempts have been made to model natural language questions as compositions of simple programs \cite{langModel2}. \cite{john} showcased the use of this approach for VQA by converting the questions into compositions of operations and using neural executors to run them. \cite{nscl} modified \cite{john}’s approach by introducing curriculum learning training objective and showed significant improvement. By the same token \cite{nsvqa} modified this further by adding structural scene representation to scene and managed to obtain really good results on the CLEVR dataset \cite{clevr}. These methods validate the symbolic reasoning approach to parsing questions.

With regards to symbolic reasoning, formal logic has traditionally been used in AI as it provides a platform for constructing rules and performing inference on symbolic data \cite{flAI}. Due to this, formal logic is highly interpretable, but not very effective when it comes to working with noisy data such as images. To circumvent this issue, neural networks are introduced to convert noisy data to a structured format, enabling the use of formal logic to develop solutions that are both robust and interpretable.

In this paper, we attempt to solve the Visual Question Answering (VQA) problem utilizing the  framework of formal logic. Inspired by the advances in scene graph generation and question to program synthesis, we propose a novel VQA pipeline consisting of: (i) conversion of scene graphs into formal logic facts, (ii) transformer-based \cite{transformer} semantic parser for translating questions into formal logic clauses and (iii) a logic inference engine performing satisfiability check on the clauses to produce answers. To elaborate, the logic clauses act as query statements on the background facts. The resulting facts and rules are highly interpretable as they are encoded as human readable formal logic statements, making it easy to analyze the reasoning process. Finally, given the rules and the facts, we utilize prolog \cite{prolog} to perform satifiability check. Importantly, the scene graph and questions are processed separately until the final stage of the pipeline, after which they are combined to obtain the answers.

Our proposed solution has the following advantanges: (i) The question to rule conversion is highly parallelized due to the use of transformers, thereby making the training of question to rule translation highly efficient on GPUs, (ii) the use of formal logic to represent images and questions makes our intermediate stages interpretable to a great extent, and can easily be analyzed by a human at every stage, (iii) it is highly data efficient with minimal change in performance when trained with fraction of the available data, refer to Section \ref{sec:clevr}. We demonstrate our proposed solution on the CLEVR dataset and the GQA dataset \cite{gqa}.    

The remainder of the paper is organized as follows: Section II details the proposed methodology and in Section III, we demonstrate the performance of our model through experiments and state the results. We then end with the discussion and conclusion in Section IV.

\section{Methodology}
\label{sec:meth}
This section details the overall methodology used to arrive at the answer given the question along with the scene information. To begin with, relevant definitions pertaining to predicate logic framework are provided followed by an overview of the overall proposed pipeline. 

\begin{figure*}[h]
\centering
\includegraphics[width=0.9\textwidth]{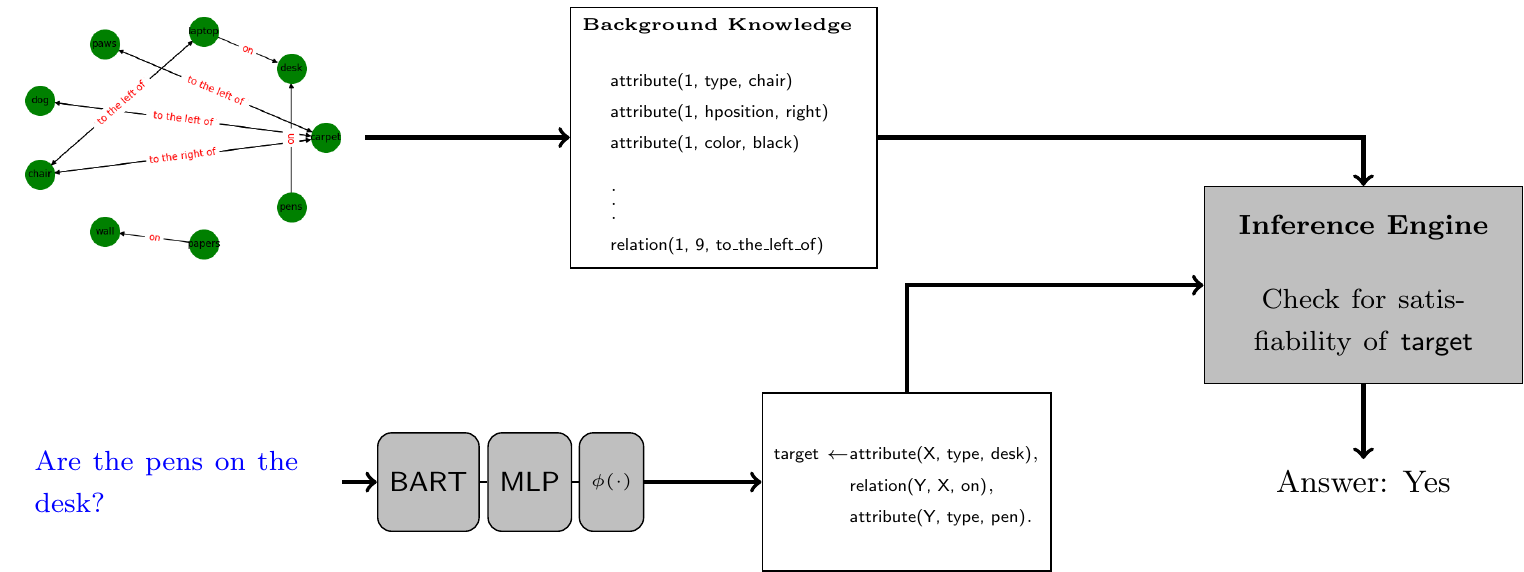}
\caption{The proposed solution has three stages: (i) scene graph to background facts conversion in the form of predicate logic, indicated by the top branch, (ii) question to target predicate logic translation using a transformer (BART \cite{bart}) based neural network and finally (iii) the logic inference which checks for the satisfiability of the final target clause. $\phi(\cdot)$ refers to the softmax function}.
\label{fig:pipeline}
\end{figure*}

\subsection{Predicate Logic.}
Predicate logic framework uses formal logic to denote facts and rules. Typically rules are expressed in the following form
\begin{equation}
a_H \gets a_1 , \ldots , a_m.
\label{eq:clause}
\end{equation}
The individual elements $a_H$ and the $a_i$'s of the rule \ref{eq:clause} are known as the atoms of the rule. Here $a_H$ is called the head of the rule and $a_1, \ldots ,a_m$ is known as the body. Rule of the above form implies that all the atoms in the body have to be true for the head rule to be true. Each atom in the rule is a n-ary Boolean function $p(x_1, \ldots, x_n)$ called a predicate, where the arguments could be variables, constants or even other predicates. A predicate expresses a relation between variables and constants in the framework. Moving forward, all constants are denoted using lower case strings/letters/numbers (e.g., \textsf{color}, \textsf{blue}, \textsf{1}, \textsf{c}) and variables are denoted using upper case letters (e.g., \textsf{W}, \textsf{X}, \textsf{Y}). A predicate is said to be grounded if all of its arguments are set to constants. Please refer to \cite{ilp} for more information on predicate logic.

\subsection{VQA Pipeline.}
Taking the scene graph and the question as the inputs, the proposed pipeline arrives at the answer by performing three separate steps, as in Figure \ref{fig:pipeline}. 

\subsubsection{Representation of Scene Graph in the form of Background facts} In this step, the given scene graph is coverted into a set of facts encoded as groundings of predicates. To that end, object IDs between $1$ to $N$ are assigned to each object in the scene, here $N$ is the total number of objects in the scene. It is important to note that these IDs are local to the scene. 
\begin{figure}[t]
\centering
\includegraphics[scale=0.7]{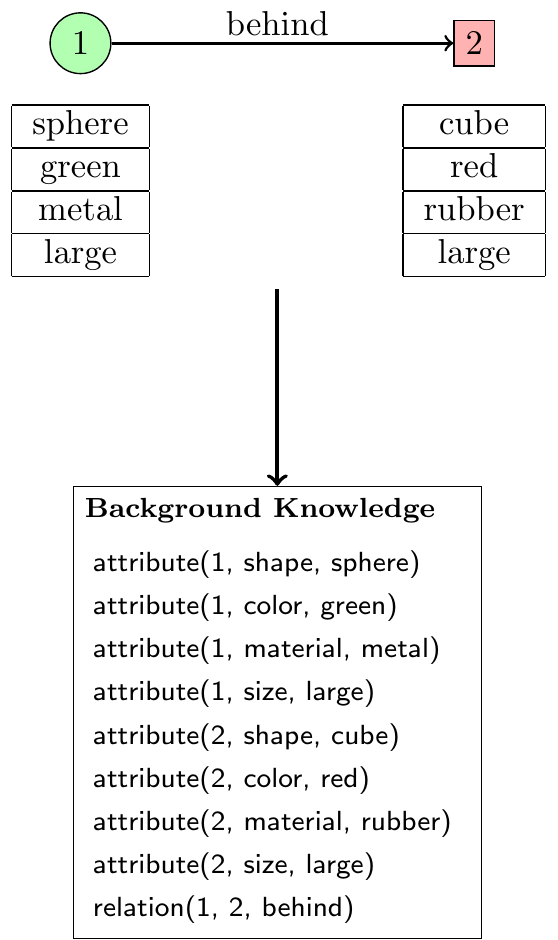}
\caption{The figure shows the conversion of scene graph to background knowledge. The scene contains two objects labelled 0 and 1. The attributes are listed in the table below the corresponding node and relation is indicated by the edge. Each attribute of the two objects are encoded as groundings of the predicate \textsf{attribute} and similarly the relation \textsf{behind} is encoded a grounding of the predicate \textsf{relation}.} 
\label{fig:img2bg}
\end{figure}

We define the following two predicates to aid in this conversion: 

\begin{enumerate}
\item \textsf{attribute}(\textsf{X, A, B}) - where the variable \textsf{X} denotes ID of the object, \textsf{A} denotes the attribute type (for example, \textsf{color}, \textsf{shape}) and \textsf{B} denotes the specific attribute (for example, \textsf{blue}, \textsf{cube}). For instance, if object 1 in the scene was blue in color then the equivalent grounding would be \textsf{attribute}(\textsf{1, color, blue}).

\item \textsf{relation}(\textsf{X, Y, A}) - where \textsf{X} and \textsf{Y} denote the object IDs, and \textsf{A} denotes the relation type (say \textsf{behind}). For instance, if objects 1 is to the left of 2 then the equivalent grounding would be \textsf{relation}(\textsf{1}, \textsf{2}, \textsf{left}).
\end{enumerate}
We then extract the object attributes and the relations from the scene graph. Iterating through every object in the scene, attributes and relations are converted to groundings of the predicates \textsf{attribute} and \textsf{relation} respectively. Finally, we obtain a list of groundings for the defined predicates which captures all the information available in the scene graph. We call this list the background knowledge corresponding to the scene. Figure \ref{fig:img2bg} shows the working for a toy example.

\subsubsection{Representation of Question in the form of first order logic rule}
\label{sec:ques2pred}

We now parse the question to generate a set of logic clauses. Given the background facts obtained from the previous step, the generated rules are such that running satisfiability check would yield in the answer. These sets of rules are formed using the predicates \textsf{attribute}, \textsf{relation} defined earlier along with a few other predicates that are necessary to capture the meaning of the question in terms of a formal logic statement. Table \ref{tab:predicates} lists the complete set of predicates used for generating the rule. It is important to note that the choice of predicates is such that it can be used with most visual question answering datasets. The constants and the groundings of the predicates would be specific to the dataset at hand. In this paper we show how the proposed framework works for the CLEVR and the GQA dataset. With slight modifications it can be incorported to work with most other VQA datasets as well. 

\begin{figure*}
\centering
\includegraphics[width=\textwidth]{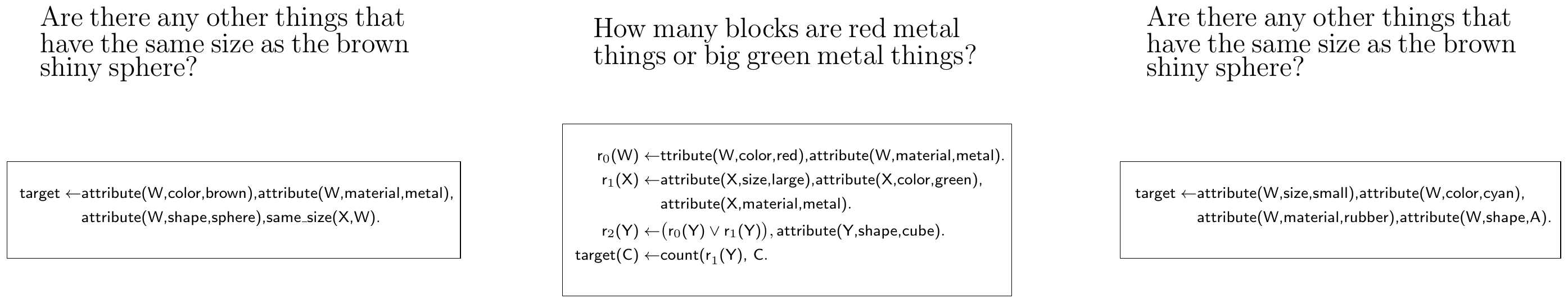}
\caption{Illustration of a few examples of questions taken from the CLEVR dataset along with the converted target rule converted by the transformer.}
\label{fig:ques2rule}
\end{figure*}

Consider the question taken from the CLEVR data set ``Are there more big green things than large purple shiny cubes?" the equivalent set of rules are shown below:
\begin{center}
\begin{minipage}{3in}
{\footnotesize
\textsf{r$_0$(W)} $\gets$ \textsf{attribute(W, size, large), attribute(W, color, green)}.\\
\textsf{r$_1$(C)} $\gets$ \textsf{count(r$_0$(W), C)}.\\ 
\textsf{r$_2$(X)} $\gets$ \textsf{attribute(X, size, large), attribute(X, color, purple), attribute(X, material, metal), attribute(X, shape, cube)}.\\
\textsf{r$_3$(C)} $\gets$ \textsf{count(r$_2$(X), C)}.\\
\textsf{target} $\gets$ \textsf{r$_1$(C$_1$), r$_3$(C$_2$)}, \textsf{greater\_than(C$_1$, C$_2$)}.
}
\end{minipage}
\end{center}
The final rule is called the target rule as it yields the required answer to the question. In the rules stated above, the arguments of the predicates contain both variables and constants. In the final stage we check to see if there are valid groundings available in the background knowledge that satisfies the rule.  Figure \ref{fig:ques2rule} shows a few examples of question target rule pair. 

To automate the process of generating the target rule given the question, we map the set of rules to a single sentence we define as the target sentence. The target sentence comprises of the predicates used in the rule along with characters to denote the conclusion of the first rule and the start of the next. For instance the target sentence for the rule mentioned previously is given by
\begin{center}
\begin{minipage}{3in}
{\footnotesize
\texttt{attribute(W, size, large),attribute(W, color, green)\textbackslash C1 \textbackslash
attribute(X, size, large), attribute(X, color, purple),
attribute(X, material, metal), attribute(X, shape, cube)\textbackslash C2 \textbackslash $>$\textbackslash} 
}
\end{minipage}
\end{center}

\begin{table}
\centering
\caption{Summary of all the predicates used in the proposed solution.}
\vspace*{0.2cm}
\begin{tabular}{|m{0.165\textwidth}|m{0.28\textwidth}|}
\hline
Predicate & \multicolumn{1}{|c|}{Definition}\\
\hline
\hline
{\small\textsf{attribute(id,t,at)}} & True if object \textsf{id} has the attribute \textsf{at} of type \textsf{t}.\\
\hline
{\small\textsf{relation(id$_1$,id$_2$,rl)}} & True if object \textsf{id$_1$} is related to object \textsf{id$_2$} by the relation \textsf{rl}.\\
\hline   
{\small\textsf{same\_size(id$_1$,id$_2$)}} & True if objects \textsf{id$_1$} and \textsf{id$_2$} have the same size attribute.\\
\hline   
{\small\textsf{same\_shape(id$_1$,id$_2$)}} & True if objects \textsf{id$_1$} and \textsf{id$_2$} have the same shape attribute.\\
\hline   
{\small\textsf{same\_color(id$_1$,id$_2$)}} & True if objects \textsf{id$_1$} and \textsf{id$_2$} have the same color attribute.\\
\hline   
{\small\textsf{same\_material(id$_1$,id$_2$)}} & True if objects \textsf{id$_1$} and \textsf{id$_2$} have the same material attribute.\\
\hline
{\small\textsf{greater\_than(n$_1$,n$_2$)}} & True if the number \textsf{n$_1$} is greater than \textsf{n$_2$}.\\
\hline
{\small\textsf{lesser\_than(n$_1$,n$_2$)}} & True if the number \textsf{n$_1$} is lesser than \textsf{n$_2$}.\\
\hline
{\small\textsf{same(c$_1$,c$_2$)}} & True if the constant \textsf{c$_1$} is same as the constant \textsf{c$_2$}.\\
\hline
{\small \textsf{count(r$_i$(X), c)}} & True if there are \textsf{c} number of solutions to \textsf{X} that satisfy \textsf{r$_i$(X)}.\\
\hline
\end{tabular}
\label{tab:predicates}
\end{table}

Given the target sentence, the rules can then be reconstructed, details regarding the construction of the target sentence and the reconstruction of the rules can be found in the supplementary materials. The rule generation can now be viewed as a machine translation problem going from english sentences to the the target sentence. To that end, we implement a transformer based sequence to sequence model to translate the question to the target sentence. We use a BART \cite{bart} pretrained on sequence regeneration task and add a MLP + softmax layer at the end. The target sentence is generated for question in the training set and the transformer  network is trained in a supervised to produce the target sentence given the question.   

\subsubsection{Logic Inference Engine.} Prolog is used to check for satisfiability of the variables present in the target rules. For binary questions (questions whose answers are either a ``yes" or a ``no") the final satisfiability of the target rule is sufficient to answer the questions. For other types of questions, we output the specific grounding that satisfies the target rule as the answer to the question.  The overall pipeline of the proposed pipeline is shown in Figure \ref{fig:pipeline}. In case the target rule produced after translation has syntax errors, then the inference engine outputs a NULL.

\section{Experiments}

In this section we show our model performance through experiments on the CLEVR and the GQA dataset.

\subsection{CLEVR Dataset} 
\label{sec:clevr}

\begin{table*}[t]
\centering
\caption{The table summarizes the resuls obtained on the CLEVR validation set. For the proposed solution we report the results when the model was trained with 10\% of training set and 100\% of the training set.}
\vspace*{0.2cm}
\begin{tabular}{|l|c|c|m{0.06\textwidth}|m{0.06\textwidth}|m{0.06\textwidth}|l|}
\hline
Models & Count & Exist & Compare Number & Compare Attribute & Query \ Attribute & Overall\\
\hline
\hline
Humans \cite{clevr} & 86.7 & 96.6 & 86.4 & 96.0 & 95.0 & 92.6\\
\hline
CNN+LSTM+SAN \cite{clevr} & 59.7 & 77.9 & 75.1 & 70.8 & 80.9 & 73.2\\
\hline
IEP \cite{john} & 92.7 & 97.1 & 98.7 & 98.9 & 98.1 & 96.9\\
\hline
NS-CL \cite{nscl} & 98.2 & 98.8 & 99.0 & 99.1 & 99.3 & 98.9\\
\hline
NS-VQA \cite{nsvqa} & 99.7 & 99.9 & 99.9 & 99.8 & 99.8 & 99.8\\
\hline
Proposed Method (10\%) & 99.3 & 99.1 & 99.1 & 99.1 & 98.7 & 99.1\\
\hline
Proposed Method (100\%) & 99.6	& 99.9 & 98.9	& 99.7	& 99.5	& 99.6\\
\hline
\end{tabular}

\label{tab:performance}
\end{table*} 

This dataset consists of rendered images containing objects of different attributes such as shape, color, size and materials. The objects are also characterized by their locations with respect to the other. These constitute the relations between the objects in the dataset, such as  left, right, front and behind. Each image has a set of questions associated with it which are generated using 90 different program templates. 

We operate under the perfect sight setting and use the scene graph provided by the dataset. For each scene in the dataset, the scene graph is converted to background knowledge as explained in Section \ref{sec:meth}. To generate the ground truth target rule, in order to train the transformer network, we make use of the functional form that the dataset provides for each question. The functional form contains a series of operations, which when performed on the image, provides the answer to the posed question. To go from the functional form to the target rule, we look at all possible operations that are a part of the functional form, and  a map is created between the operations and the elements of the target rule.

The transformer model is trained with a variable length question as the input and the target sentence (defined in section \ref{sec:meth}) as the output. In particular, we use a pretrained BART model for the transformer, in conjunction with a MLP and softmax layer. The final softmax layer outputs the necessary tokens to generate the target sentence. The use of transformer makes the training highly parallelizable. We train the model using Google Colab on a NVIDIA Tesla V100 GPU. The model was trained for a total of 4 epochs on the training set using ADAM \cite{adam} optimizer with a learning rate of $10^{-4}$.

At this point, we have the scenes converted into background knowledge and the questions translated to logic rules, which act as queries on the background knowledge. Then, prolog is used to run a satisfiability check on the rules to obtain the answer. 

The proposed method was evaluated on the validation set and we obtained a near perfect accuracy of 99.6\%, close to the existing state of the art \cite{nsvqa} (99.8\%). This shows that the proposed framework of using formal logic is capable of solving the task of visual question answering. Table \ref{tab:performance} summarizes the performance along with a comparison with the baselines. 

\vspace*{-3pt}

To further test the learning capabilities of our proposed method, we train our translation model on 10\% of the questions from the training set (62,997 questions) equally distributed between the 90 question templates. Even with a fraction of training samples our proposed method attains 99.16\% accuracy on the validation set. We chose the validation set over the test set to test our model as the scene graphs are readily available in the validation set. For the purpose of training the transformer, we split the training data into training plus validation to monitor the training process. 

Figure \ref{fig:ques2rule} shows a few examples of the questions and the corresponding target rule produced by the translator for the CLEVR dataset. From knowing the definition of the predicates (defined in Section \ref{sec:ques2pred}), the target rule can easily be interpreted by a human. For example, considering the first example in Figure \ref{fig:ques2rule} we need to check if there are any other things with the same size as the brown shiny sphere. Looking at the corresponding target clause we see that it is a conjunction of four predicates where the first three predicates filter out objects with the attributes \textit{brown, metal, sphere} and the last predicate checks if there are two objects with the same set of attributes mentioned earlier. Therefore looking at the translated rules we know exactly what is happening underneath and what the satisfiability check is looking for.

\subsection{CLEVR CoGenT.} CLEVR cogent was proposed by \cite{clevr} to test the generalizing capabilities of VQA models when the dataset is biased, refer to \cite{clevr} for more information on the construction of the dataset. 

We train our model on split A using the same training procedure as the CLEVR dataset and test on the validation sets of both the splits. Our proposed method again achieves identical near perfect validation accuracy of 99.52\% on both the splits. The independent treatment of images and language information makes it easy for the model to learn separate disentangled representations for the individual attributes as evident from the final accuracy. As seen from Table \ref{tab:cogent}, the proposed model performs as well as the state of the art NS-VQA with a significant reduction in the training burden. 

\begin{table}
\centering
\caption{Performance comparison of the proposed method on the CLEVR CoGenT dataset when trained on split A.}
\vspace*{0.2cm}
\begin{tabular}{|c|c|c|}
\hline
Models & Split A & Split B\\
\hline
\hline
NS-CL \cite{nscl} & 98.8\% & 98.9\%\\
\hline
NS-VQA \cite{nsvqa} & 99.8 & 99.7 \\
\hline
Proposed Method & 99.5 & 99.5\\
\hline
\end{tabular}
\label{tab:cogent}
\end{table}

\subsection{GQA Dataset.} GQA is another visual question answering dataset proposed by \cite{gqa}, where questions are constructed using compositions of a set of operations. We follow the same pipeline used in the CLEVR dataset. Again, we operate under pure sight condition and use the scene graphs provided by the dataset for conversion into background knowledge. 

\begin{table}
\centering
\caption{Performance of the proposed method on the GQA validation set. For the proposed solution we report results when the model was tested on full validation set (FV) and the reduced validation set. (RV)}
\vspace*{0.2cm}
\begin{tabular}{|c|c|}
\hline
Method & Accuracy\\
\hline
\hline
GN+MAC \cite{sgVqa2} & 96.3\\
\hline
Proposed Method-FV & 74.3\\
\hline
Proposed Method-RV & 93.1\\
\hline
\end{tabular}
\label{tab:gqa}
\end{table}

To train the translation model, we make use of the functional form provided by the dataset for each question to produce the target sentence. Similar to CLEVR, every operation in the functional form is mapped to a predicate/rule in the target sentence. We found the functional forms provided by the dataset to be inconsistent in the semantics used for defining the arguments associated with the operations in the functional form, refer to the supplementary materials for more details. This resulted in incorrect target sentences for certain questions, therefore we restrict our training to only those questions having consistent functional forms.

On the GQA validation set, we obtained an overall accuracy of (74\%) but this contains several questions corresponding to inconsistent functional forms for which the model wasn't trained for. When the validation set is limited to those questions which have a consistent functional form, then the validation accuracy becomes (93\%), which is a better indicator of the model's performance on the GQA datasets. This is comparable to the current state of the art for the GQA dataset under perfect site configuration \cite{sgVqa2} (96.3\%). Aside from the fact that the proposed solution is also highly interpretable and the reasoning steps are easy to follow, the training of the transformer is much more easier to perform than a Graph Neural Network used in \cite{sgVqa2}. Table \ref{tab:gqa} summarizes the model performance on the GQA dataset. Again, since the scene graphs for the test set are not readily provided by the dataset we only evaluate the model on the validation set as we operate in the perfect sight setting. 

\section{Discussion and Conclusion}
In this paper, we showcased the feasibility of using an interpretable formal logic based approach to solve the visual question answering problem. We circumvent the inflexibility of formal logic systems to noisy inputs by using transformers to translate natural language questions to interpretable logic rules. From the conducted experiments, it can be seen that our proposed solution is competitive and even manages to achieve near perfect accuracy on the CLEVR dataset. It is also data efficient and there is only a slight drop in accuracy even when trained with just 10\% data from the CLEVR dataset.

\bibliographystyle{IEEEtran}
\bibliography{references.bib}

\end{document}